\pdfoutput=1

\documentclass[11pt]{article}

\usepackage{eacl2023}

\usepackage{times}
\usepackage{latexsym}
\usepackage[T1]{fontenc}
\usepackage[utf8]{inputenc}
\usepackage{microtype}
\usepackage{inconsolata}
\usepackage{graphicx}
\usepackage{enumitem}
\usepackage{algorithm}
\usepackage{algpseudocode}
\usepackage[normalem]{ulem}
\useunder{\uline}{\ul}{}

\newcommand{\commentout}[1]{}

\usepackage{enumitem}

\usepackage{todonotes}

\title{Evaluating the Effectiveness of Data Augmentation for Emotion Classification in Low-Resource Settings}

\author{Aashish Arora \and Elsbeth Turcan \\
Columbia University \\
New York, NY, USA\\
\texttt{\{aa4830,ect2150\}@columbia.edu} \\}

\begin{document}
\maketitle

\begin{abstract}
Data augmentation has the potential to improve the performance of machine learning models by increasing the amount of training data available. In this study, we evaluated the effectiveness of different data augmentation techniques for a multi-label emotion classification task using a low-resource dataset. Our results showed that Back Translation outperformed autoencoder-based approaches and that generating multiple examples per training instance led to further performance improvement. In addition, we found that Back Translation generated the most diverse set of unigrams and trigrams. These findings demonstrate the utility of Back Translation in enhancing the performance of emotion classification models in resource-limited situations.
\end{abstract}

\section{Introduction}
\label{sec:intro}
The proliferation of unlabeled data has the potential to significantly enhance the performance of machine learning models. However, the lack of labelled data can present a challenge in utilizing this resource, particularly in low-resource languages and domains. 

One approach to addressing this issue is the use of pseudo labelling, which involves using a model trained on labelled data to annotate a portion of the unlabeled data \cite{imagenet-classification}, \cite{10.5555/3454287.3455291}. This annotated data can be used in conjunction with the original labelled data to improve the performance and robustness of the model on downstream tasks.

An additional technique for improving model performance is knowledge distillation \cite{44873}, which involves transferring the knowledge contained in a larger, more complex model (the "teacher") to a smaller, simpler model (the "student"). By distilling the knowledge from the teacher model into the student model, it is possible to achieve better performance on a task with a smaller model, which can be especially beneficial in low-resource settings where computational resources may be limited. 

Data augmentation is another strategy that can be employed to improve model performance, particularly in the context of small or low-resource datasets. This involves generating additional data samples from existing ones in order to increase the diversity and quantity of the dataset. This can help to improve generalization and reduce the risk of overfitting. Within the field of natural language processing (NLP), data augmentation has gained significant popularity as a means of addressing the often small and unbalanced nature of real-world datasets. 

There are several methods of data augmentation in NLP, including synonym replacement, random insertion, and random deletion \cite{EDA}. These techniques allow for the generation of new sentences or paragraphs that are similar in meaning to the original data but differ in their wording and structure. These techniques gave a better performance on five benchmark text classification tasks \cite{EDA}. Back translation, which involves translating text from one language to another and then back to the original language is also used as an augmentation technique \cite{sennrich-etal-2016-improving}.

Another approach to data augmentation in NLP is to use pre-trained transformer models \cite{attention}. Transformer-based models, such as BERT \cite{bert}, GPT-3 \cite{gpt} and BART \cite{bart}, are large language models (LLM) that have been trained on massive amounts of data and can perform various NLP tasks, including sequence classification, language translation, summarization, and question answering. By fine-tuning these models on a specific task, it is possible to generate high-quality and diverse data samples that can be used to improve the performance of downstream tasks, especially in scenarios where the dataset is small or low-resourced. 

This study investigates the impact of different data augmentation methods on a new corpus. Specifically, 
\begin{enumerate}
    \item Utilize a state-of-the-art multi-label emotion classification model to annotate a large dataset with pseudo labels.
    \item Downsample the dataset to simulate a low-resource scenario.
    \item Employ various generative models to synthesize additional data with predetermined labels.
    \item Evaluate the diversity and class-label preservation of the synthetic data. 
    \item Use the resulting low-resource dataset in conjunction with the augmented data to train a model and assess its performance on the new corpus.
\end{enumerate}

\begin{figure}
    \centering
    \includegraphics[scale=0.5]
{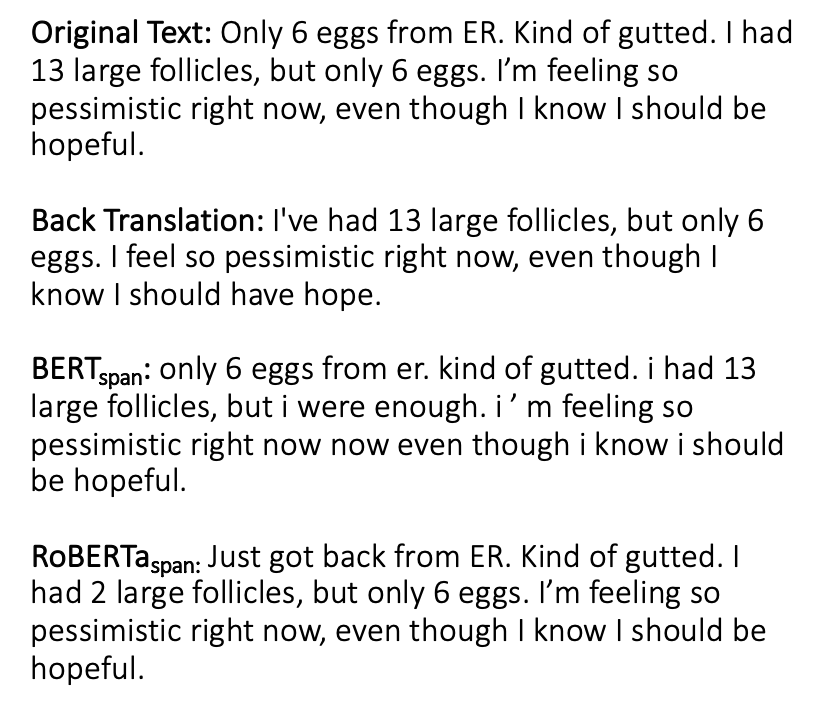}
    \caption{Illustration of generated examples using Back Translation, BERT, and RoBERTa based data augmentation techniques. Back Translation produced the most diverse set of unigrams and trigrams, while the autoencoder-based methods (BERT and RoBERTa) modified existing words or spans. The original sentence is displayed for comparison.}
    \label{fig:data_diversity}
\end{figure}

\section{Related Work}
\label{sec:relwork}
\paragraph{Emotion Classification} Emotions are central to language and thought, being both familiar and nuanced. In the SemEval-2018 Task 1: Affect in Tweets \cite{semeval-2018}, a series of subtasks were proposed to infer the affective emotions of individuals from their tweets, using labelled data in English, Arabic, and Spanish. These subtasks included emotion intensity regression, emotion intensity ordinal classification, valence regression, valence ordinal classification, and emotion classification. Specifically, the emotion classification subtask involves classifying tweets as "neutral or no emotion" or as one or more of 11 predefined emotions that best reflect the mental state of the tweet's author. In our study, we aim to use the emotion classification subtask dataset to classify Reddit posts written in English as "neutral or no emotion" or expressing one or more of 11 predefined emotions. These classifications will be used as pseudo labels for a large, unlabeled Reddit dataset.

In the SemEval-2018 Task 1: Affect in Tweets \cite{semeval-2018} competition, the authors of the NTUA \cite{ntua} paper employed a Bi-LSTM model with a multi-layer self-attention mechanism for the affect classification of English tweets. This model made use of word2vec embeddings and affective word features, and utilized transfer learning with pretraining on the SemEval-2017 Task 4: Sentiment Analysis in Twitter \cite{semeval-2017} dataset in order to address the limited amount of task-specific training data. This approach achieved a ranking of first place in the "Multi-Label Emotion Classification" subtask of SemEval-2018 \cite{semeval-2018}. In the current study, we will utilize this model, referred to as the NLP-SLP model in this paper, to conduct pseudo-labelling on a Reddit dataset and perform multi-label emotion classification on the new corpus to assess the effectiveness of various data augmentation techniques.

BERTweet \cite{bertweet}, a large-scale pre-trained language model for English tweets with an architecture similar to $\textnormal{BERT}_{base}$ \cite{bert} and trained using the RoBERTa \cite{roberta} pre-training procedure, has been demonstrated to outperform previous state-of-the-art models on three Tweet NLP tasks: Part-Of-Speech tagging, Named-Entity Recognition, and Text Classification. However, when this model was applied to the task of multi-label emotion classification \cite{semeval-2018}, the results were not promising. Furthermore, the reported results for 3-class sentiment analysis on the SemEval2017 Task 4A \cite{semeval-2017} dataset could not be reproduced. Therefore, this reported state-of-the-art model for various NLP tasks in the social media domain is not used in our study.

\paragraph{Data Augmentation} In \cite{mental-health}, the authors explored the use of different data augmentation techniques to generate additional text for mental health classification using three specific methods: Easy Data Augmentation \cite{EDA}, conditional BERT \cite{cbert}, and Back Translation. These techniques were applied to two publicly available social media datasets \cite{turcan-mckeown-2019-dreaddit}, \cite{09427} and the resulting augmented data was used to train three different classifiers (Random Forest, Support Vector Machine, and Logistic Regression). The results showed that the classifiers' performance significantly improved when trained on the augmented data.

\cite{da} studied the use of pre-trained transformer models, including auto-regressive models (GPT-2 \cite{gpt2}), auto-encoder models (BERT \cite{bert}), and seq2seq models (BART \cite{bart}), for data augmentation in natural language processing (NLP) tasks. In their work, they proposed a method for conditioning the pre-trained models for data augmentation by prepending class labels to text sequences. They compared the performance of these models on classification benchmarks in a low-resource setting. They also examined how the different data augmentation methods using pre-trained models differ in terms of data diversity and how well they preserve class-label information.  

The authors \cite{gal} propose the "generate, annotate, and learn" (GAL) framework for generating high-quality, task-specific text. To generate the text, they either fine-tune large language models (LLMs) on inputs from the relevant task or prompt LLMs with a few examples. The generated text is then annotated with soft pseudo labels using the best available classifier, and the annotated text is used for knowledge distillation and self-training. The authors show that training new supervised models on a combination of labelled and pseudo-labelled data leads to significant improvements across various applications. They also present theoretical and empirical arguments against using class-conditional LLMs to generate synthetic labelled text instead of unlabeled text.

\section{Data}
\subsection{IESO dataset}
\label{sec:ieso_data}
\begin{figure}
    \centering
    \includegraphics[scale=0.5]{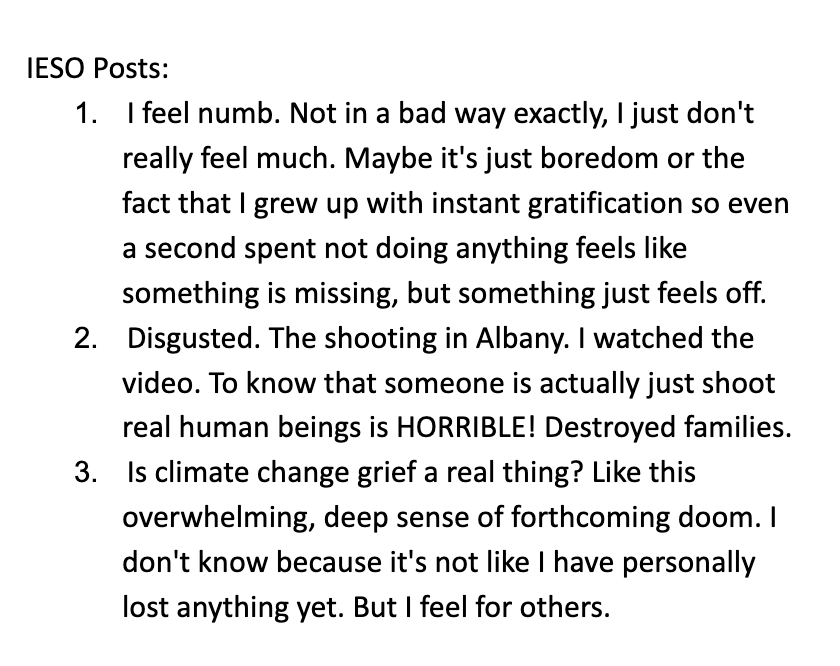}
    \caption{Examples of posts in the IESO dataset.}
    \label{fig:ieso_post}
\end{figure}

In this study, we utilize the IESO dataset, a collection of self-reported emotional experiences expressed in written posts and rated on a 1-10 scale. The dataset includes 877 entries, in which individuals described their emotions and potentially the events that led to those emotions. A range of emotions was recorded, including anger, sadness, stress, happiness, lonely, calmness, excitement, anxiety, annoyance, hope, despair, guilty, afraid, disgust, surprise, and numbness. Some individuals also reported not experiencing any strong emotions. All of the posts were made by individuals recruited for the study. Few examples of IESO posts are shown in figure \ref{fig:ieso_post}.

\subsection{Reddit Dataset}
\label{sec:reddit_data}
\begin{figure}
    \centering
    \includegraphics[scale=0.5]{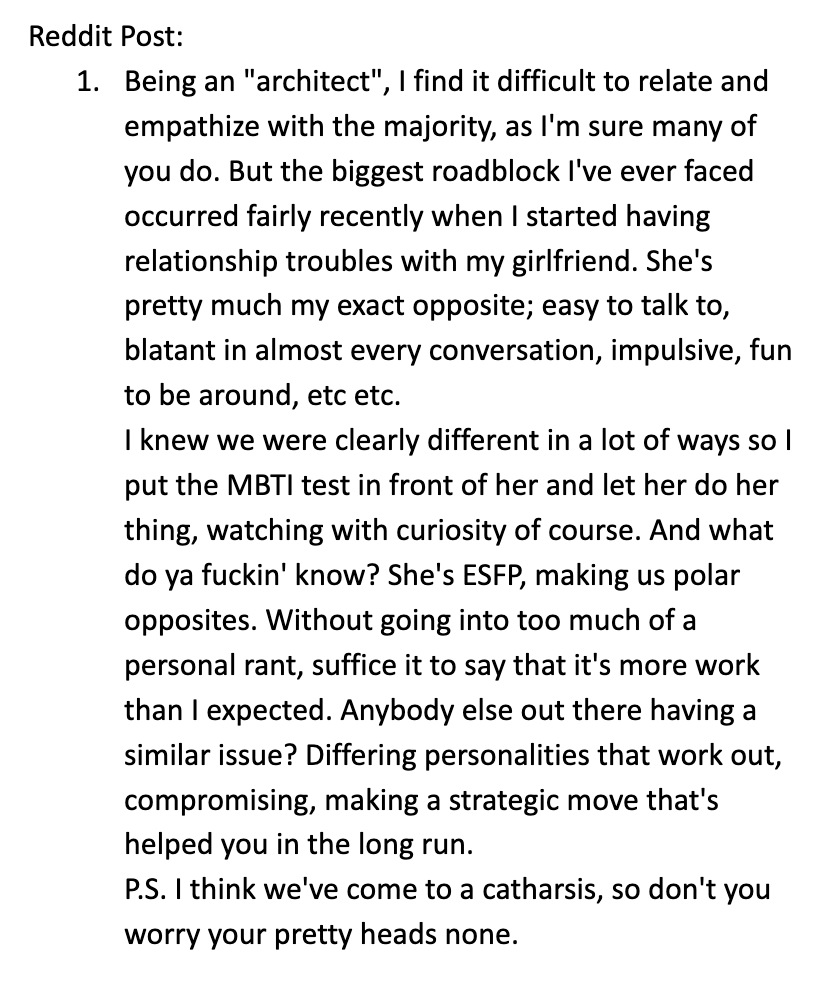}
    \caption{Example of post in the Reddit dataset}
    \label{fig:reddit_post}
\end{figure}

In addition, we obtained a large dataset of Reddit posts from various subreddits, including a title and potentially additional text for each post, as well as the subreddit to which it belongs. The dataset includes a "grief" field indicating whether the subreddit on which the post was made belongs to one of the subreddits catering to people expressing grief, and an "after\_first\_grief" field indicating whether the post is a follow-up related to grief on a previous post. However, we do not use these fields in our study as it is possible for posts expressing grief to appear in subreddits not specifically intended for expressing grief, and vice versa. As a result, this dataset is not labelled for any specific emotions. The dataset consists of 224,493 posts collected through web crawling. An example of Reddit post is shown in figure \ref{fig:reddit_post}

\section{Methodology}
In this study, we utilized Back Translation and auto-encoder models such as BERT \cite{bert} and RoBERTa \cite{roberta} to synthesize data, and then employed the NTUA-SLP model \cite{ntua} to annotate the synthesized data. The annotated data is combined with low-resource labelled data and used to fine-tune the same classifier. The findings of this research support the hypothesis that generating high-quality data can enhance the performance and robustness of a model in low-resource settings.

\subsection{Data Preprocessing}
\label{sec:data_preprocessing}
As mentioned in Section \ref{sec:ieso_data}, the IESO dataset consists of descriptions of individuals' emotional states and the events that may have caused those emotions. For the purposes of this study, we have combined these two descriptions into a single entity, referred to as an "IESO post". In addition, we transformed the self-reported emotional experiences of users, rated on a scale of 1-10, into a binary format using a threshold of 4. Ratings lower than 4 were converted to 0, indicating that the corresponding post does not express the relevant emotion, while ratings of 4 or higher were converted to 1. As previously mentioned in Section \ref{sec:ieso_data}, the IESO dataset includes annotations for a range of emotions. However, this set of emotions does not align with the emotions covered in the SemEval-2018 Task 1 \cite{semeval-2018} dataset, which was used to train the NTUA-SLP model \cite{ntua} and which we sought to utilize as a pre-trained model for multi-label emotion classification on the IESO dataset. To reconcile this difference, we established a mapping between the two datasets, as illustrated in Table \ref{mapping}. Any emotions present in the IESO dataset but not included in this mapping were excluded from the scope of this study.

\begin{table}[]
\begin{tabular}{|c|c|}
\hline
\textbf{IESO Emotions} & \textbf{SemEval-2018 Emotions} \\ \hline
angry                  & anger                          \\ \hline
excited                & anticipation                   \\ \hline
disgusted              & disgust                        \\ \hline
fear                   & afraid                         \\ \hline
happy                  & joy                            \\ \hline
hopeful                & optimism                       \\ \hline
despaired              & pessimism                      \\ \hline
sad                    & sadness                        \\ \hline
surprised              & surprise                       \\ \hline
\end{tabular}
\caption{Table presenting the mapping between the emotions annotated in the IESO dataset and those covered in the SemEval-2018 Task 1 \cite{semeval-2018}. The emotions annotated in the IESO dataset but not included in this mapping and therefore excluded from the scope of this study are: disapprove, stressed, lonely, calm, anxious, annoyed, guilty, numb, no\_strong\_emotion.}
\label{mapping}
\end{table}

To facilitate our analysis, we have combined the title and text of each Reddit post, if available, into a single entity referred to as a "Reddit post" for the purpose of this study.

\subsection{Data Augmentation}
\begin{algorithm}
    \caption{Data Augmentation Approach}
    \begin{algorithmic}
        \item \textbf{Input}: A pre-trained model $G$ and a training dataset $D_{train}$
        \item \textbf{Output}: A synthetic dataset $D_{synthetic}$  \\
        \State Fine-tune $G$ using $D_{train}$ to obtain $G_{tuned}$
        \State $D_{synthetic}$ = \{\}
        \For{$x_i \in D_{tuned}$}
            \State Synthesize $s$ examples using $G_{tuned}$ 
            \State Add these examples in $D_{synthetic}$
        \EndFor
    \end{algorithmic}
\end{algorithm}

In accordance with the methodology described in \cite{da}, Algorithm 1 outlines the data augmentation process utilized in this study. Given a pre-trained model $G$, a training dataset $D_{train} = \{x_i, y_i\}^1_n$, where $x_i = \{w_j\}^i_k$ is a sequence of k tokens, $y_i$ is the label, we want to synthesize additional data $D_{synthetic}$. As part of this study, we will generate $s=\{1,3,5\}$ examples for each example in the training dataset. The effectiveness of these multi-fold data augmentation techniques will be evaluated as part of this study.

\subsubsection{Data Augmentation using Auto Encoder Language Models}
\label{sec:auto_encoder}
To augment the data using auto-encoder models such as BERT \cite{bert} and RoBERTa \cite{roberta}, we employed the Mask Language Modelling (MLM) objective \cite{bert}, which randomly masks a portion of the tokens in a sentence with a special "mask" token and attempts to predict the original token from the masked sentence. In order to synthesize data expressing various emotions, we modified this approach by masking out tokens that convey strong negative emotions instead of selecting them randomly. To identify these tokens, we compiled a list of words with high arousal and low valence scores \cite{vad} and calculated the cosine similarity \cite{cosine} of each non-functional word in a Reddit post with the list of strong negative emotional (SNE) words using Glove embeddings \cite{glove} as the word representation. If there are no words in a sentence that directly match the list of SNE words, this process allows us to identify semantically similar words for masking in the Mask Language Modelling process. Words with higher cosine similarity scores to any of the strong negative emotional words are given more weight in the masking process.

We attempted three techniques to generate a masked sentence. 
\begin{enumerate}
    \item $AE_{token}$: randomly masking out 15\% of the tokens in the sentence based on their cosine similarity to the list of SNE words.
    \item $AE_{span}$: masking out a continuous chunk of 5 tokens containing the randomly chosen word to be masked.
    \item $AE_{constituent}$: masking out the entire constituent of the randomly chosen word to be masked.
\end{enumerate}
For the purposes of this study, AE refers to auto-encoder models such as BERT and RoBERTa.

For the $AE_{constituent}$ technique, we utilized the spaCy library's \cite{spacy} implementation to identify the constituent of a token. However, this approach was not successful because the tokenization of the auto-encoder models and the spaCy library differ, and we were unable to align them due to the introduction of special characters by the auto-encoder model's tokenizers, which modifies the original sentence.

\subsubsection{Data Augmentation using Back Translation}
\label{sec:back_translation}
To augment the training data using Back Translation, we used French as an intermediate language since research has shown that translations between English and French tend to produce the highest quality translations \cite{en-fr}. Each Reddit post was converted into a sentence in French and then back into English, resulting in a new synthesized post that was added to the dataset. The huggingface implementation of MarianMT \cite{mariannmt} was utilized for the Back Translation process.

\subsection{Simulating a low-resource data setting}
\label{sec:low_resource}
To exclude posts from our Reddit dataset that are either clearly or borderline conveying a particular emotion, we utilize the confidence score output by the NTUA-SLP model \cite{ntua} on a Reddit post for each of the 11 emotions \cite{semeval-2018}. The NTUA-SLP model applies a hard sigmoid (as described in equation \ref{sec:hard_sigmoid}) function to this confidence score to predict whether a particular emotion is being conveyed by the Reddit post. To exclude posts that are either too easy or too difficult for the model to predict as conveying a particular emotion, we require that the confidence score for a predicted emotion falls within the range of 0.3 to 0.8 on a scale of 0 to 1. This approach results in a dataset that includes high-quality and diverse data samples. 

\begin{equation}
\label{sec:hard_sigmoid}
    f(x) = max(0, min(1, \frac{x+1}{2}))
\end{equation}

In order to simulate a low-resource data setting, we randomly select 1000 examples from the downsampled annotated Reddit dataset, of which we use 700 for training and 300 for validation purposes for the different data augmentation techniques.

\subsection{Evaluation}
\label{sec:evaluation}
As previously discussed in \cite{da}, we conduct both intrinsic and extrinsic evaluations to assess the effectiveness of different data augmentation techniques. For extrinsic evaluation, we incorporate the generated examples into the low-resource training data and evaluate the performance on the full IESO dataset. 

For intrinsic evaluation, we consider two aspects of the generated text: the preservation of meaning and class information of the input sentence, as measured by fine-tuning the NTUA-SLP \cite{ntua} model on low-resource training data and testing the synthesized data using the fine-tuned classifier, and the diversity of the generated output, as measured by the type-token ratio \cite{type-token} calculated by dividing the number of unique n-grams by the total number of n-grams in the generated text.

\section{Experimental Setup}
\subsection{Auto-Encoder based Data Augmentation}
In this study, we employed the $BERT_{base}$ \cite{bert} and $\textnormal{RoBERTa}_{base}$ \cite{roberta} implementations from huggingface to perform data augmentation using an auto-encoder approach. To ensure compatibility with our model, we enforced a maximum sequence length of 512 tokens for each Reddit post and discarded any posts exceeding this length. We optimized the auto-encoder-based data augmentation model using Adam \cite{adam} optimization with a batch size of 32 across 2 V100-SXM2 GPUs and a fixed learning rate of $5e^{-5}$. We fine-tuned both $\textnormal{BERT}_{base}$ and $\textnormal{RoBERTa}_{base}$ for 25 epochs and used the model with the lowest loss on the validation dataset for synthesizing new data.

\subsection{Back Translation based Data Augmentation}
For the purpose of back translation, we utilized Marian \cite{mariannmt}, a neural machine translation framework that includes an automatic differentiation engine and is implemented in C++. We employed pre-trained English-French and French-English translation models from Marian.

\subsection{Multi-Label Emotion Classification}
The architecture and training and testing procedure for our emotion classification model are identical to that of the NTUA-SLP model \cite{ntua}. The model employs a Bi-LSTM architecture with a multi-layer self-attention mechanism. For more information, please refer to \cite{ntua}.

\section{Results and Discussion}
\subsection{Pre-trained Model Comparison}
\subsubsection{Classification Performance}
As previously mentioned, we combined the generated examples with the low-resource training data to fine-tune the classifier and evaluated the performance on the full IESO dataset. To obtain a reliable baseline result on the IESO dataset, we utilized the NTUA-SLP model \cite{ntua} without any fine-tuning. The results, presented in Table \ref{result-emotions}, demonstrate that Back Translation consistently outperformed the baseline and other data augmentation techniques. Additionally, the results in Table \ref{result-emotions} indicate that multi-folding the generated data (i.e. generating more than one example per training instance) significantly improved performance compared to only doubling the dataset by synthesizing one example per training example.

\begin{table}
\centering
\resizebox{\columnwidth}{!}{%
\begin{tabular}{lccc} 
\hline
\multicolumn{1}{c}{} & \multicolumn{1}{c}{\begin{tabular}[c]{@{}c@{}}\textbf{Jaccard similarity }\\\textbf{coefficient}\end{tabular}} & \multicolumn{1}{l}{\textbf{F1-Macro}} & \textbf{F1-Micro}  \\ 
\hline
Baseline                                        & 0.254                                                                                                                                & 0.36                                                        & 0.258                     \\
BT                                              & 0.271                                                                                                                                & 0.371                                                      & 0.269  \\ 
$\textnormal{BT}_{-1}$ & \textbf{0.287}                                                                                                                                & 0.374                                                      & 0.269                    \\ 
$\textnormal{BERT}_{token}$                                     & 0.259                                                                                                                                & 0.364                                                      & 0.259                    \\ 
$\textnormal{BERT}_{token-1}$                                  & 0.262                                                                                                                                & 0.372                                                      & 0.269                    \\ 
$\textnormal{BERT}_{token-3}$                                  & 0.265                                                                                                                                & 0.375                                                       & 0.271                     \\
$\textnormal{BERT}_{token-5}$                                  & 0.269                                                                                                                                & 0.379                                                       & 0.270                     \\
$\textnormal{BERT}_{span}$                                      & 0.262                                                                                                                                & 0.365                                                      & 0.261                    \\
$\textnormal{BERT}_{span-1}$                                   & 0.281                                                                                                                                & 0.380                                                      & 0.270                    \\
$\textnormal{BERT}_{span-3}$                                   & 0.283                                                                                                                                & 0.383                                                       & 0.273                     \\
$\textnormal{BERT}_{span-5}$                                   & 0.283                                                                                                                                & \textbf{0.386}                                                       & 0.270                     \\
$\textnormal{RoBERTa}_{token}$                                  & 0.263                                                                                                                                & 0.362                                                      & 0.261                    \\
$\textnormal{RoBERTa}_{token-1}$                               & 0.268                                                                                                                                & 0.373                                                       & 0.264                     \\
$\textnormal{RoBERTa}_{token-3}$                               & 0.273                                                                                                                                & 0.374                                                       & 0.262                     \\
$\textnormal{RoBERTa}_{token-5}$                               & 0.277                                                                                                                                & 0.375                                                       & \textbf{0.272}                     \\
$\textnormal{RoBERTa}_{span}$                                   & 0.264                                                                                                                                & 0.369                                                      & 0.259                    \\
$\textnormal{RoBERTa}_{span-1}$                                & 0.271                                                                                                                                & 0.371                                                       & 0.261                     \\
$\textnormal{RoBERTa}_{span-3}$                                & 0.273                                                                                                                                & 0.377                                                       & 0.265                     \\
$\textnormal{RoBERTa}_{span-5}$                                & 0.275                                                                                                                                & 0.378                                                       & 0.262                     \\
\hline
\end{tabular}%
}
\caption{Extrinsic evaluations of different data aug-
mentation techniques. The NTUA-SLP \cite{ntua} model was used to produce baseline results by evaluating it directly on the IESO dataset without any fine-tuning. The variable $x$ in $\textnormal{AE}_{tech-x}$ denotes the number of examples generated from a single training instance, where $tech \in $ \{token, span\} and AE $\in $ \{BERT, RoBERTa\}.}
\label{result-emotions}
\end{table}

\subsubsection{Generated Data Semantic Fidelity}
\begin{figure}
    \centering
    \includegraphics[scale=0.4]{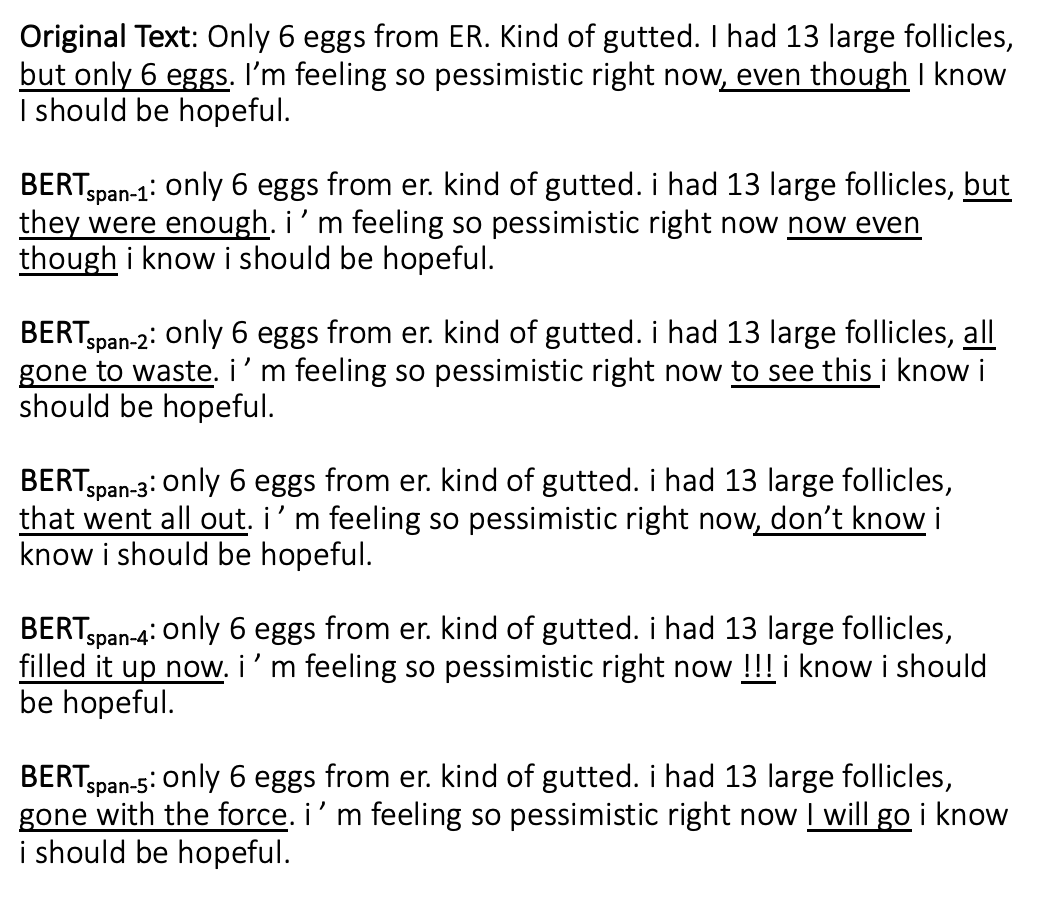}
    \caption{Figure illustrating the impact of multi-fold generation using BERT-based data augmentation methods on semantic fidelity. The masked tokens in the original sentence are indicated with underlining. As the number of synthesized sentences per training example increases, we see a decline in the ability of the generated text to accurately convey the meaning of the original sentence.}
    \label{fig:semantic_fidelity}
\end{figure}

As mentioned in Section \ref{sec:evaluation}, we used the NLP-SLP \cite{ntua} classifier to evaluate the semantic fidelity of the generated text. The results, presented in the table \ref{result-fidelity}, show that $\textnormal{BERT}_{span}$ outperformed Back Translation and other auto-encoder-based methods. This demonstrates the effectiveness of state-of-the-art masked language models in preserving the semantics of the language. Additionally, we found that generating multiple examples per training example resulted in a reduction in the ability of the generated text to retain the original class information, as shown in Figure \ref{fig:semantic_fidelity}. This suggests that using a greedy decoding approach to predict the most probable replacement for a mask token may not effectively preserve the content of the original sentence.

\begin{table}
\centering
\resizebox{\columnwidth}{!}{%
\begin{tabular}{lccc} 
\hline
\multicolumn{1}{c}{} & \multicolumn{1}{c}{\begin{tabular}[c]{@{}c@{}}\textbf{Jaccard similarity }\\\textbf{coefficient}\end{tabular}} & \multicolumn{1}{l}{\textbf{F1-Macro}} & \textbf{F1-Micro}  \\ 
\hline
BT                    & 0.83                                                                                                            & 0.658                                  & 0.874              \\ 
$\textnormal{BERT}_{token-1}$                  & 0.814                                                                                                           & 0.75                                   & 0.8556             \\ 
$\textnormal{BERT}_{token-3}$        & 0.80                                                                                                            & 0.73                                   & 0.8552             \\
$\textnormal{BERT}_{token-5}$        & 0.793                                                                                                           & 0.71                                   & 0.834              \\
$\textnormal{BERT}_{span-1}$         & \textbf{0.89}                                                                                                   & 0.896                                  & \textbf{0.92}      \\
$\textnormal{BERT}_{span-3}$         & 0.87                                                                                                            & \textbf{0.898}                         & 0.89               \\
$\textnormal{BERT}_{span-5}$         & 0.83                                                                                                            & 0.885                                  & 0.83               \\
$\textnormal{RoBERTa}_{token-1}$      & 0.82                                                                                                            & 0.83                                   & 0.86               \\
$\textnormal{RoBERTa}_{token-3}$     & 0.84                                                                                                            & 0.85                                   & 0.87               \\
$\textnormal{RoBERTa}_{token-5}$     & 0.81                                                                                                            & 0.823                                  & 0.854              \\
$\textnormal{RoBERTa}_{span-1}$      & 0.872                                                                                                           & 0.788                                  & 0.902              \\
$\textnormal{RoBERTa}_{span-3}$      & 0.871                                                                                                           & 0.792                                  & 0.91               \\
$\textnormal{RoBERTa}_{span-5}$      & 0.865                                                                                                           & 0.773                                  & 0.82               \\
\hline
\end{tabular}%
}
\caption{Semantic fidelity scores for different data augmentation techniques. The variable $x$ in $\textnormal{AE}_{tech-x}$ denotes the number of examples generated from a single training instance, where $tech \in $ \{token, span\} and AE $\in $ \{BERT, RoBERTa\}.}
\label{result-fidelity}
\end{table}

\subsubsection{Generated Data Diversity}
To provide further insight into the generated data, we examined the type-token ratio. The results, shown in the table \ref{result-diversity}, indicate that Back Translation generated the most diverse set of unigrams and trigrams. Unlike auto-encoder methods which modify existing words or spans, Back Translation synthesizes a new sequence of tokens that aims to preserve the semantic content of the input sentence, which often results in the introduction of previously unseen unigrams and trigrams. This can be seen in Figure \ref{fig:data_diversity}, which displays the original sentence and the synthesized data produced using various data augmentation techniques.
\begin{table}[]
\begin{tabular}{lcc}
\hline
\multicolumn{3}{c}{Token-Type Ratio}                                                     \\ \hline
\multicolumn{1}{c}{\textbf{}} & \multicolumn{1}{c}{\textbf{Uni-gram}} & Tri-gram       \\ \hline
BT                             & \textbf{0.718}                         & \textbf{0.980} \\
$\textnormal{BERT}_{token}$                    & 0.66                                   & 0.977          \\
$\textnormal{BERT}_{span}$                     & 0.647                                  & 0.979          \\
$\textnormal{RoBERTa}_{token}$                 & 0.674                                  & ,0.978         \\
$\textnormal{RoBERTa}_{span}$                  & 0.656                         & 0.978          \\ \hline
\end{tabular}
\caption{Table summarizing the data diversity of various data augmentation techniques, as measured by the type-token ratio.}
\label{result-diversity}
\end{table}

\section{Limitation}
In this study, we apply various data augmentation techniques in a low-resource setting and demonstrate improvements in the performance and robustness of a classifier. However, we find that generating more than one augmented example per original training example may lead to a loss of information conveyed in the original text, as discussed in Section. 

The NTUA-SLP model \cite{ntua}, which has been trained and tested on a large amount of Twitter data, is being applied to generate pseudo-labels on a large unlabeled Reddit dataset. However, Twitter tweets are significantly shorter in length and structured differently than Reddit posts, and there is no publicly available, reliable annotated Reddit dataset for emotion classification that we can use to evaluate the model's performance on Reddit data. As a result, it is possible that the pseudo-labels produced for the Reddit dataset may not accurately reflect the actual annotations.

Additionally, the IESO data includes a range of emotions that do not map directly to those in the SemEval-2018 dataset \cite{semeval-2018}, resulting in examples in our IESO dataset that may not convey any emotions covered in the SemEval-2018 dataset. This may be misinterpreted as the post not conveying any emotions, which may not be accurate.

\section{Future Work}
As a future direction, it would be interesting to investigate the effectiveness of auto-regressor models such as GPT-3 \cite{gpt} and seq2seq models like BART \cite{bart} for data generation and their impact on emotion classification performance in low-resource settings. However, due to budget and time constraints, we were unable to include these models in our study.

Longformer \cite{longformer}, a transformer-based model that utilizes an attention mechanism that scales linearly with sequence length, is proposed as a solution to accurately represent long Reddit posts, which may be challenging for pre-trained models such as BERT and RNN-based models like NTUA due to their limited maximum sequence length. Previous research \cite{long-range} has shown that Longformer outperforms vanilla transformer models on text with long sequences. It would be interesting to evaluate the performance of different data augmentation techniques on longer sequences of text using Longformers. 

Additionally, the IESO dataset, which includes descriptions of individuals' emotional states and events that may have caused those emotions, can be used as a gold-labelled rational-detection dataset. By applying different data augmentation techniques to synthesize additional data from a limited amount of labelled data, we can train a robust rational-detection model.

\bibliography{custom}

\begin{thebibliography}{31}
\expandafter\ifx\csname natexlab\endcsname\relax\def\natexlab#1{#1}\fi

\bibitem[{Ansari et~al.(2021)Ansari, Garg, and Saxena}]{mental-health}
Gunjan Ansari, Muskan Garg, and Chandni Saxena. 2021.
\newblock \href {https://doi.org/10.48550/ARXIV.2112.10064} {Data augmentation
  for mental health classification on social media}.

\bibitem[{Baziotis et~al.(2018)Baziotis, Athanasiou, Chronopoulou, Kolovou,
  Paraskevopoulos, Ellinas, Narayanan, and Potamianos}]{ntua}
Christos Baziotis, Nikos Athanasiou, Alexandra Chronopoulou, Athanasia Kolovou,
  Georgios Paraskevopoulos, Nikolaos Ellinas, Shrikanth Narayanan, and
  Alexandros Potamianos. 2018.
\newblock \href {https://doi.org/10.48550/ARXIV.1804.06658} {Ntua-slp at
  semeval-2018 task 1: Predicting affective content in tweets with deep
  attentive rnns and transfer learning}.

\bibitem[{Beltagy et~al.(2020)Beltagy, Peters, and Cohan}]{longformer}
Iz~Beltagy, Matthew~E. Peters, and Arman Cohan. 2020.
\newblock \href {https://doi.org/10.48550/ARXIV.2004.05150} {Longformer: The
  long-document transformer}.

\bibitem[{Brown et~al.(2020)Brown, Mann, Ryder, Subbiah, Kaplan, Dhariwal,
  Neelakantan, Shyam, Sastry, Askell, Agarwal, Herbert-Voss, Krueger, Henighan,
  Child, Ramesh, Ziegler, Wu, Winter, Hesse, Chen, Sigler, Litwin, Gray, Chess,
  Clark, Berner, McCandlish, Radford, Sutskever, and Amodei}]{gpt}
Tom~B. Brown, Benjamin Mann, Nick Ryder, Melanie Subbiah, Jared Kaplan,
  Prafulla Dhariwal, Arvind Neelakantan, Pranav Shyam, Girish Sastry, Amanda
  Askell, Sandhini Agarwal, Ariel Herbert-Voss, Gretchen Krueger, Tom Henighan,
  Rewon Child, Aditya Ramesh, Daniel~M. Ziegler, Jeffrey Wu, Clemens Winter,
  Christopher Hesse, Mark Chen, Eric Sigler, Mateusz Litwin, Scott Gray,
  Benjamin Chess, Jack Clark, Christopher Berner, Sam McCandlish, Alec Radford,
  Ilya Sutskever, and Dario Amodei. 2020.
\newblock \href {https://doi.org/10.48550/ARXIV.2005.14165} {Language models
  are few-shot learners}.

\bibitem[{Carmon et~al.(2019)Carmon, Raghunathan, Schmidt, Liang, and
  Duchi}]{10.5555/3454287.3455291}
Yair Carmon, Aditi Raghunathan, Ludwig Schmidt, Percy Liang, and John~C. Duchi.
  2019.
\newblock \emph{Unlabeled Data Improves Adversarial Robustness}. Curran
  Associates Inc., Red Hook, NY, USA.

\bibitem[{Devlin et~al.(2018)Devlin, Chang, Lee, and Toutanova}]{bert}
Jacob Devlin, Ming-Wei Chang, Kenton Lee, and Kristina Toutanova. 2018.
\newblock \href {https://doi.org/10.48550/ARXIV.1810.04805} {Bert: Pre-training
  of deep bidirectional transformers for language understanding}.

\bibitem[{Haque et~al.(2021)Haque, Reddi, and Giallanza}]{09427}
Ayaan Haque, Viraaj Reddi, and Tyler Giallanza. 2021.
\newblock \href {https://doi.org/10.48550/ARXIV.2102.09427} {Deep learning for
  suicide and depression identification with unsupervised label correction}.

\bibitem[{He et~al.(2021)He, Nassar, Kiros, Haffari, and Norouzi}]{gal}
Xuanli He, Islam Nassar, Jamie Kiros, Gholamreza Haffari, and Mohammad Norouzi.
  2021.
\newblock \href {https://doi.org/10.48550/ARXIV.2106.06168} {Generate,
  annotate, and learn: Nlp with synthetic text}.

\bibitem[{Hinton et~al.(2015)Hinton, Vinyals, and Dean}]{44873}
Geoffrey Hinton, Oriol Vinyals, and Jeffrey Dean. 2015.
\newblock \href {http://arxiv.org/abs/1503.02531} {Distilling the knowledge in
  a neural network}.
\newblock In \emph{NIPS Deep Learning and Representation Learning Workshop}.

\bibitem[{Honnibal and Montani(2017)}]{spacy}
Matthew Honnibal and Ines Montani. 2017.
\newblock {spaCy 2}: Natural language understanding with {B}loom embeddings,
  convolutional neural networks and incremental parsing.
\newblock To appear.

\bibitem[{Junczys-Dowmunt et~al.(2018)Junczys-Dowmunt, Grundkiewicz, Dwojak,
  Hoang, Heafield, Neckermann, Seide, Germann, Aji, Bogoychev, Martins, and
  Birch}]{mariannmt}
Marcin Junczys-Dowmunt, Roman Grundkiewicz, Tomasz Dwojak, Hieu Hoang, Kenneth
  Heafield, Tom Neckermann, Frank Seide, Ulrich Germann, Alham~Fikri Aji,
  Nikolay Bogoychev, André F.~T. Martins, and Alexandra Birch. 2018.
\newblock \href {https://doi.org/10.48550/ARXIV.1804.00344} {Marian: Fast
  neural machine translation in c++}.

\bibitem[{Kingma and Ba(2014)}]{adam}
Diederik~P. Kingma and Jimmy Ba. 2014.
\newblock \href {https://doi.org/10.48550/ARXIV.1412.6980} {Adam: A method for
  stochastic optimization}.

\bibitem[{Kumar et~al.(2020)Kumar, Choudhary, and Cho}]{da}
Varun Kumar, Ashutosh Choudhary, and Eunah Cho. 2020.
\newblock \href {https://doi.org/10.48550/ARXIV.2003.02245} {Data augmentation
  using pre-trained transformer models}.

\bibitem[{Lewis et~al.(2019)Lewis, Liu, Goyal, Ghazvininejad, Mohamed, Levy,
  Stoyanov, and Zettlemoyer}]{bart}
Mike Lewis, Yinhan Liu, Naman Goyal, Marjan Ghazvininejad, Abdelrahman Mohamed,
  Omer Levy, Ves Stoyanov, and Luke Zettlemoyer. 2019.
\newblock \href {https://doi.org/10.48550/ARXIV.1910.13461} {Bart: Denoising
  sequence-to-sequence pre-training for natural language generation,
  translation, and comprehension}.

\bibitem[{Liu et~al.(2019)Liu, Ott, Goyal, Du, Joshi, Chen, Levy, Lewis,
  Zettlemoyer, and Stoyanov}]{roberta}
Yinhan Liu, Myle Ott, Naman Goyal, Jingfei Du, Mandar Joshi, Danqi Chen, Omer
  Levy, Mike Lewis, Luke Zettlemoyer, and Veselin Stoyanov. 2019.
\newblock \href {https://doi.org/10.48550/ARXIV.1907.11692} {Roberta: A
  robustly optimized bert pretraining approach}.

\bibitem[{Mikolov et~al.(2013)Mikolov, Sutskever, Chen, Corrado, and
  Dean}]{cosine}
Tomas Mikolov, Ilya Sutskever, Kai Chen, Greg Corrado, and Jeffrey Dean. 2013.
\newblock \href {https://doi.org/10.48550/ARXIV.1310.4546} {Distributed
  representations of words and phrases and their compositionality}.

\bibitem[{Mohammad(2018)}]{vad}
Saif Mohammad. 2018.
\newblock \href {https://doi.org/10.18653/v1/P18-1017} {Obtaining reliable
  human ratings of valence, arousal, and dominance for 20,000 {E}nglish words}.
\newblock In \emph{Proceedings of the 56th Annual Meeting of the Association
  for Computational Linguistics (Volume 1: Long Papers)}, pages 174--184,
  Melbourne, Australia. Association for Computational Linguistics.

\bibitem[{Mohammad et~al.(2018)Mohammad, Bravo-Marquez, Salameh, and
  Kiritchenko}]{semeval-2018}
Saif Mohammad, Felipe Bravo-Marquez, Mohammad Salameh, and Svetlana
  Kiritchenko. 2018.
\newblock \href {https://doi.org/10.18653/v1/S18-1001} {{S}em{E}val-2018 task
  1: Affect in tweets}.
\newblock In \emph{Proceedings of the 12th International Workshop on Semantic
  Evaluation}, pages 1--17, New Orleans, Louisiana. Association for
  Computational Linguistics.

\bibitem[{Nguyen et~al.(2020)Nguyen, Vu, and Tuan~Nguyen}]{bertweet}
Dat~Quoc Nguyen, Thanh Vu, and Anh Tuan~Nguyen. 2020.
\newblock \href {https://doi.org/10.18653/v1/2020.emnlp-demos.2} {{BERT}weet: A
  pre-trained language model for {E}nglish tweets}.
\newblock In \emph{Proceedings of the 2020 Conference on Empirical Methods in
  Natural Language Processing: System Demonstrations}, pages 9--14, Online.
  Association for Computational Linguistics.

\bibitem[{Pennington et~al.(2014)Pennington, Socher, and Manning}]{glove}
Jeffrey Pennington, Richard Socher, and Christopher Manning. 2014.
\newblock \href {https://doi.org/10.3115/v1/D14-1162} {{G}lo{V}e: Global
  vectors for word representation}.
\newblock In \emph{Proceedings of the 2014 Conference on Empirical Methods in
  Natural Language Processing ({EMNLP})}, pages 1532--1543, Doha, Qatar.
  Association for Computational Linguistics.

\bibitem[{Prabhumoye et~al.(2018)Prabhumoye, Tsvetkov, Salakhutdinov, and
  Black}]{en-fr}
Shrimai Prabhumoye, Yulia Tsvetkov, Ruslan Salakhutdinov, and Alan~W Black.
  2018.
\newblock \href {https://doi.org/10.48550/ARXIV.1804.09000} {Style transfer
  through back-translation}.

\bibitem[{Radford et~al.(2019)Radford, Wu, Child, Luan, Amodei, and
  Sutskever}]{gpt2}
Alec Radford, Jeff Wu, Rewon Child, David Luan, Dario Amodei, and Ilya
  Sutskever. 2019.
\newblock Language models are unsupervised multitask learners.

\bibitem[{Roemmele et~al.(2017)Roemmele, Gordon, and Swanson}]{type-token}
Melissa Roemmele, Andrew~S. Gordon, and Reid Swanson. 2017.
\newblock Evaluating story generation systems using automated linguistic
  analyses.

\bibitem[{Rosenthal et~al.(2017)Rosenthal, Farra, and Nakov}]{semeval-2017}
Sara Rosenthal, Noura Farra, and Preslav Nakov. 2017.
\newblock \href {https://doi.org/10.18653/v1/S17-2088} {{S}em{E}val-2017 task
  4: Sentiment analysis in {T}witter}.
\newblock In \emph{Proceedings of the 11th International Workshop on Semantic
  Evaluation ({S}em{E}val-2017)}, pages 502--518, Vancouver, Canada.
  Association for Computational Linguistics.

\bibitem[{Sennrich et~al.(2016)Sennrich, Haddow, and
  Birch}]{sennrich-etal-2016-improving}
Rico Sennrich, Barry Haddow, and Alexandra Birch. 2016.
\newblock \href {https://doi.org/10.18653/v1/P16-1009} {Improving neural
  machine translation models with monolingual data}.
\newblock In \emph{Proceedings of the 54th Annual Meeting of the Association
  for Computational Linguistics (Volume 1: Long Papers)}, pages 86--96, Berlin,
  Germany. Association for Computational Linguistics.

\bibitem[{Tay et~al.(2020)Tay, Dehghani, Abnar, Shen, Bahri, Pham, Rao, Yang,
  Ruder, and Metzler}]{long-range}
Yi~Tay, Mostafa Dehghani, Samira Abnar, Yikang Shen, Dara Bahri, Philip Pham,
  Jinfeng Rao, Liu Yang, Sebastian Ruder, and Donald Metzler. 2020.
\newblock \href {https://doi.org/10.48550/ARXIV.2011.04006} {Long range arena:
  A benchmark for efficient transformers}.

\bibitem[{Turcan and McKeown(2019)}]{turcan-mckeown-2019-dreaddit}
Elsbeth Turcan and Kathy McKeown. 2019.
\newblock \href {https://doi.org/10.18653/v1/D19-6213} {{D}readdit: A {R}eddit
  dataset for stress analysis in social media}.
\newblock In \emph{Proceedings of the Tenth International Workshop on Health
  Text Mining and Information Analysis (LOUHI 2019)}, pages 97--107, Hong Kong.
  Association for Computational Linguistics.

\bibitem[{Vaswani et~al.(2017)Vaswani, Shazeer, Parmar, Uszkoreit, Jones,
  Gomez, Kaiser, and Polosukhin}]{attention}
Ashish Vaswani, Noam Shazeer, Niki Parmar, Jakob Uszkoreit, Llion Jones,
  Aidan~N. Gomez, Lukasz Kaiser, and Illia Polosukhin. 2017.
\newblock \href {https://doi.org/10.48550/ARXIV.1706.03762} {Attention is all
  you need}.

\bibitem[{Wei and Zou(2019)}]{EDA}
Jason Wei and Kai Zou. 2019.
\newblock \href {https://doi.org/10.48550/ARXIV.1901.11196} {Eda: Easy data
  augmentation techniques for boosting performance on text classification
  tasks}.

\bibitem[{Wu et~al.(2018)Wu, Lv, Zang, Han, and Hu}]{cbert}
Xing Wu, Shangwen Lv, Liangjun Zang, Jizhong Han, and Songlin Hu. 2018.
\newblock \href {https://doi.org/10.48550/ARXIV.1812.06705} {Conditional bert
  contextual augmentation}.

\bibitem[{Xie et~al.(2020)Xie, Luong, Hovy, and Le}]{imagenet-classification}
Qizhe Xie, Minh-Thang Luong, Eduard Hovy, and Quoc~V. Le. 2020.
\newblock \href {https://doi.org/10.1109/CVPR42600.2020.01070} {Self-training
  with noisy student improves imagenet classification}.
\newblock In \emph{2020 IEEE/CVF Conference on Computer Vision and Pattern
  Recognition (CVPR)}, pages 10684--10695.

\end{thebibliography}
\bibliographystyle{acl_natbib}
\appendix

\end{document}